\documentclass{article}

\usepackage{PRIMEarxiv}

\usepackage[utf8]{inputenc} 
\usepackage[T1]{fontenc}    
\usepackage{hyperref}       
\usepackage{url}            
\usepackage{booktabs}       
\usepackage{amsfonts}       
\usepackage{nicefrac}       
\usepackage{microtype}      
\usepackage{lipsum}
\usepackage[numbers,sort&compress]{natbib}
\usepackage{tabularx}
\usepackage{fancyhdr}       
\usepackage{graphicx}       
\usepackage{amsmath}
\usepackage{threeparttable}
\graphicspath{{media/}}     

\title{ScVLM: Enhancing Vision-Language Model for Safety-Critical Event Understanding} 

\author{
  Liang Shi\thanks{First author.} \\
  Department of Statistics, \\
  Virginia Tech Transportation Institute, \\
  Virginia Polytechnic Institute and State University; \\
  \texttt{sliang@vt.edu} \\
  \And
  Boyu Jiang\\
  Department of Statistics, \\
  Virginia Polytechnic Institute and State University, \\
  \texttt{Boyuj@vt.edu} \\
  \And
  Tong Zeng\\
  Department of Computer Science, \\
  Virginia Polytechnic Institute and State University, \\
  \texttt{tongzeng@vt.edu} \\
  \And
  Feng Guo\thanks{Corresponding author.} \\
  Department of Statistics, \\
  Virginia Tech Transportation Institute, \\
  Virginia Polytechnic Institute and State University; \\
  \texttt{feng.guo@vt.edu} \\
}

\hypersetup{
    pdfinfo={
        Title={ScVLM: Enhancing Vision-Language Model for Safety-Critical Event Understanding},
        Author={Liang Shi, Boyu Jiang, Tong Zeng, Feng Guo},
        Keywords={Driving Safety-Critical Events, Vision-Language Models, Supervised Learning, Contrastive Learning, Event Description Rationality},
        Journal-ref={Proceedings of the IEEE/CVF Winter Conference on Applications of Computer Vision (WACV), 2025}
    }
}

\begin{document}

\maketitle

\begin{abstract}
Accurately identifying, understanding and describing traffic safety-critical events (SCEs), including crashes, tire strikes, and near-crashes, is crucial for advanced driver assistance systems, automated driving systems, and traffic safety. As SCEs are rare events, most general vision-language models (VLMs) have not been trained sufficiently to link SCE videos and narratives, which could lead to hallucinations and missing key safety characteristics. Here, we introduce ScVLM, a novel hybrid methodology that integrates \textbf{s}upervised and \textbf{c}ontrastive learning techniques to classify the severity and types of SCEs, as well as to generate narrative descriptions of SCEs. This approach utilizes classification to enhance VLMs' comprehension of driving videos and improve the rationality of event descriptions. The proposed approach is trained on and evaluated by more than 8,600 SCEs from the Second Strategic Highway Research Program Naturalistic Driving Study dataset, the largest publicly accessible driving dataset with videos and SCE annotations. The results demonstrate the superiority of the proposed approach in generating contextually accurate event descriptions and mitigating VLM hallucinations. The code will be available at \href{https://github.com/datadrivenwheels/ScVLM}{https://github.com/datadrivenwheels/ScVLM}
\end{abstract}


\section{Introduction}

In the domain of traffic safety and automatic driving, vision language models (VLMs) have demonstrated strong and robust capabilities in perception, scene understanding, decision-making, and adaptability to novel scenarios \cite{gan2024drive, Shoman_2024_CVPR, tian2024drivevlm, zhou2024vision}. VLMs can proficiently interpret environmental information surrounding the vehicle and possess foundational insights into traffic accidents and potential risk factors \cite{Shoman_2024_CVPR, zhou2024vision, llvmad2025position}. However, despite these advances, challenges still exist in accurately identifying safety-critical events (SCEs), including crashes and near-crashes. Furthermore, understanding the nature of these SCEs, such as conflicts with a lead vehicle, remains elusive. This information is crucial for assessing driving safety.

\begin{figure}
  \centering
  \includegraphics[scale=0.4]{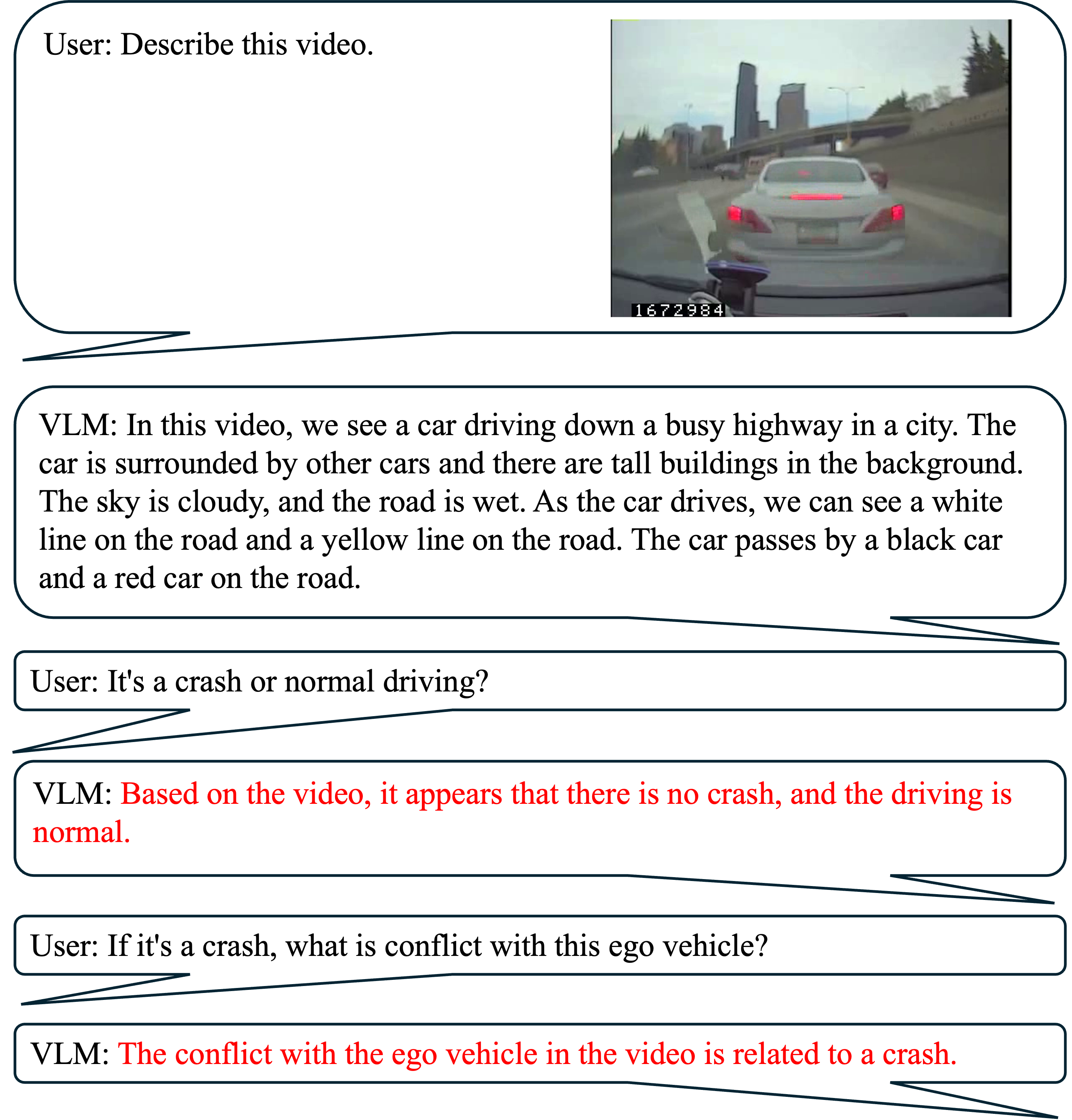}
  \caption{Example scene understanding result by VideoLLaMA2 (red highlights are the incorrect answers).}
  \label{fig:llama_example}
\end{figure}


Figure \ref{fig:llama_example} illustrates the capabilities of an advanced VLM,
VideoLLaMA 2 \cite{damonlpsg2024videollama2}, 
in understanding SCEs. This model exhibits an excellent understanding of static environmental contexts, including weather conditions and the immediate surroundings. However, its ability to discern dynamic elements crucial for SCE analysis—such as distinguishing between crash and normal driving scenarios or identifying the nature of conflicts (e.g., with a leading or parked vehicle)—is still constrained. These findings underscore the necessity for improved models capable of more effectively interpreting dynamic information in SCE videos.

The limitations observed in VideoLLaMA 2's performance on SCE analysis can be attributed to two key factors. (1) The rarity of SCEs in real-world scenarios results in insufficient training data for general VLMs to establish connections between SCE videos and corresponding narratives \cite{shi2022real, cassese2023assessing}. (2) The scarcity of relevant training examples can lead to hallucinations and the omission of crucial safety characteristics in the model's interpretations. Additionally, the abstract nature of event types and conflict types poses a significant challenge for VLMs to accurately identify scenarios \cite{gendron2024large}.

This work introduces a novel hybrid approach for generating narratives from driving videos, with a focus on SCEs. The approach combines supervised learning, contrastive learning, and language models to provide accurate and coherent event descriptions. Supervised learning is employed for event type identification (i.e., crashes, tire strikes, near-crashes, normal driving), taking advantage of its effectiveness for task-specific classification. For conflict type identification, contrastive learning is used to capture semantic dependencies between labels and rich textual information. To interpret environmental context, a VLM is utilized to accurately recognize concrete objects within the video. Finally, a Large Language Model (LLM) integrates the outputs from the supervised and contrastive learning components, along with the environmental context, to generate coherent event narratives.

The primary contribution of this study is the development of an accurate event description generator that addresses the issue of hallucinations in VLMs. 
The proposed approach enhances prediction precision for these elements, thereby guiding the VLM to generate more accurate event descriptions.

The evaluation of the proposed approach utilized data from the Second Strategic Highway Research Program (SHRP 2) Naturalistic Driving Study (NDS), which is the largest publicly accessible NDS dataset to date, containing over 1 million hours of continuous driving data \cite{hankey2016description}.  The SHRP 2 NDS data includes rich driving information from multiple cameras, kinematic sensors, radar, and GPS. From the continuous driving data, a dedicated project was conducted to identify SCEs and randomly selected normal driving baselines \cite{hankey2016description}, including four distinct event types: crashes, tire strikes, near-crashes, and normal driving baselines. SCEs went through a rigorous data annotation process to extract the nature of the conflict. The annotations provide detailed conflict type labels for SCEs, covering scenarios like conflicts with a lead vehicle, single-vehicle conflicts, and conflicts with a vehicle turning into another's path in the same direction. This rich dataset is ideal for evaluating the effectiveness of the proposed hybrid approach.

\section{Related Works}

\noindent \textbf{VLM for Driving Scene Understanding} VLMs combine visual and language processing to interpret driving scenarios and aid decision-making. DriveVLM incorporates reasoning modules for scene description and analysis, addressing spatial reasoning and computational challenges by proposing a hybrid system that combines VLMs with traditional autonomous driving pipelines \cite{tian2024drivevlm}. DriveScenify utilizes advanced VLMs to generate contextually relevant responses based on driving scene videos, aiming to enhance urban mobility and road safety \cite{drivescenify2023multimodal}. \citet{Shoman_2024_CVPR} propose a parallel architecture that integrates object detection, tracking, and natural language generation to produce detailed descriptions of traffic events, thereby improving traffic safety through comprehensive event analysis. \citet{jain2024semantic} integrate VLMs with multi-sensor data to enhance the comprehension of traffic dynamics and interactions among road users and infrastructure.

While research on VLMs for general scene understanding in driving contexts has expanded significantly, the specific focus on SCEs, which are vital for improving safety and reliability in autonomous vehicles, remains under-explored. Even if some works mention crashes or traffic accidents \cite{Shoman_2024_CVPR, zhou2024vision, tian2024drivevlm}, they do not explore the intricacies of these events in depth.

\vspace{5pt}

\noindent \textbf{Supervised Learning and Contrastive Learning} Supervised learning and contrastive learning are two popular approaches for driving video scene classification tasks \cite{taccari2018classification, Shi_2024_CVPR, shi2024two, yang2023accident, chen2023clip2scene}. Supervised learning relies on one-hot or figure-coded labels to train models \cite{yang2024comprehensive}, while contrastive learning, particularly in a video-text manner, takes advantage of the relationships between different modals of the data to learn useful representations \cite{xu2021videoclipcontrastivepretrainingzeroshot}.  In supervised learning, state-of-the-art and efficient methods such as SlowFast \cite{Feichtenhofer_2019_ICCV}, Swin Transformer \cite{liu2022video}, and TimeSformer \cite{bertasius2021spacetimeattentionneedvideo} have proven effective for video scene understanding. In contrastive learning, inspired by CLIP \cite{radford2021learning}, notable approaches like X-CLIP \cite{ni2022expanding} and ActionCLIP \cite{wang2023actionclip}  excel in video understanding, particularly in few-shot tasks. X-CLIP introduced a lightweight cross-frame attention mechanism and proposed a video-adaptive textual prompting scheme to handle video-text datasets \cite{ni2022expanding}. ActionCLIP introduced textual and visual adapters to enhance the model's ability to process and understand text and video modalities \cite{wang2023actionclip}.

\section{VLM-based Driving SCE Analysis}

\begin{figure*}[h!]
  \centering
  \includegraphics[scale=0.22]{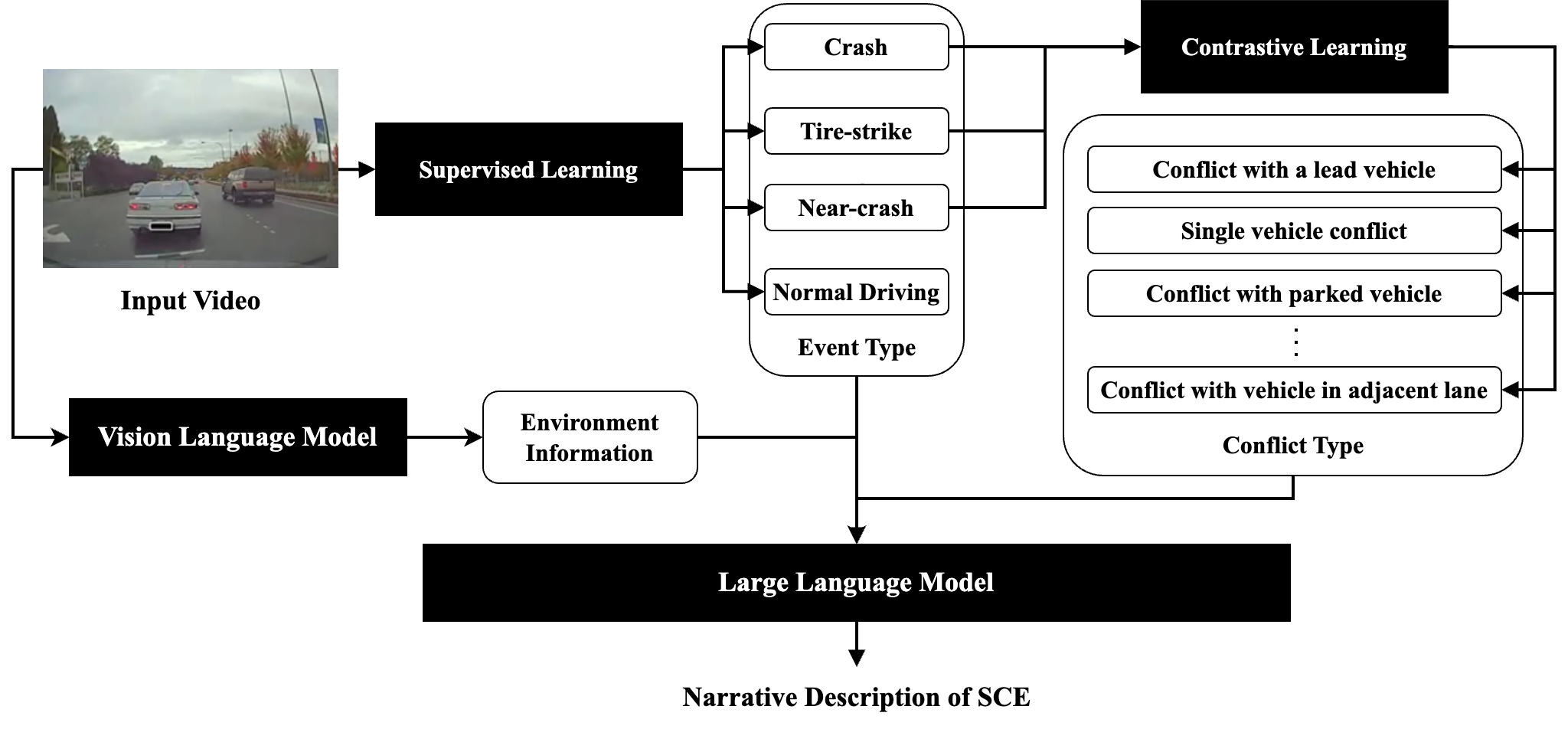}
  \caption {\label{fig:model_process}The proposed multi-stage approach for generating narrative descriptions of SCEs from driving videos. The process integrates supervised learning for event classification (e.g., crash, near-crash) and contrastive learning for conflict type identification (e.g., conflict with lead vehicle, single vehicle conflict). The VLM extracts visual and environmental information, which is further refined by an LLM to produce a detailed narrative of the SCE.}
\end{figure*}

The proposed approach for generating comprehensive and accurate descriptions of SCEs comprises four distinct stages, as depicted in Figure \ref{fig:model_process}. Initially, a general VLM is employed to extract environmental information from event videos. Subsequently, a supervised learning approach classifies front-view video into four categories: crashes, tire strikes, near-crashes, and normal driving. The third stage utilizes a contrastive learning approach to identify 16 distinct conflict types, such as conflict with a lead vehicle, parked vehicle, and following vehicle. Finally, the framework integrates event classification, conflict type, and environmental context into an LLM to synthesize a comprehensive event description.

\subsection{Supervised Learning for Event Type Classification}

\begin{figure}
  \centering
  \includegraphics[scale=0.45]{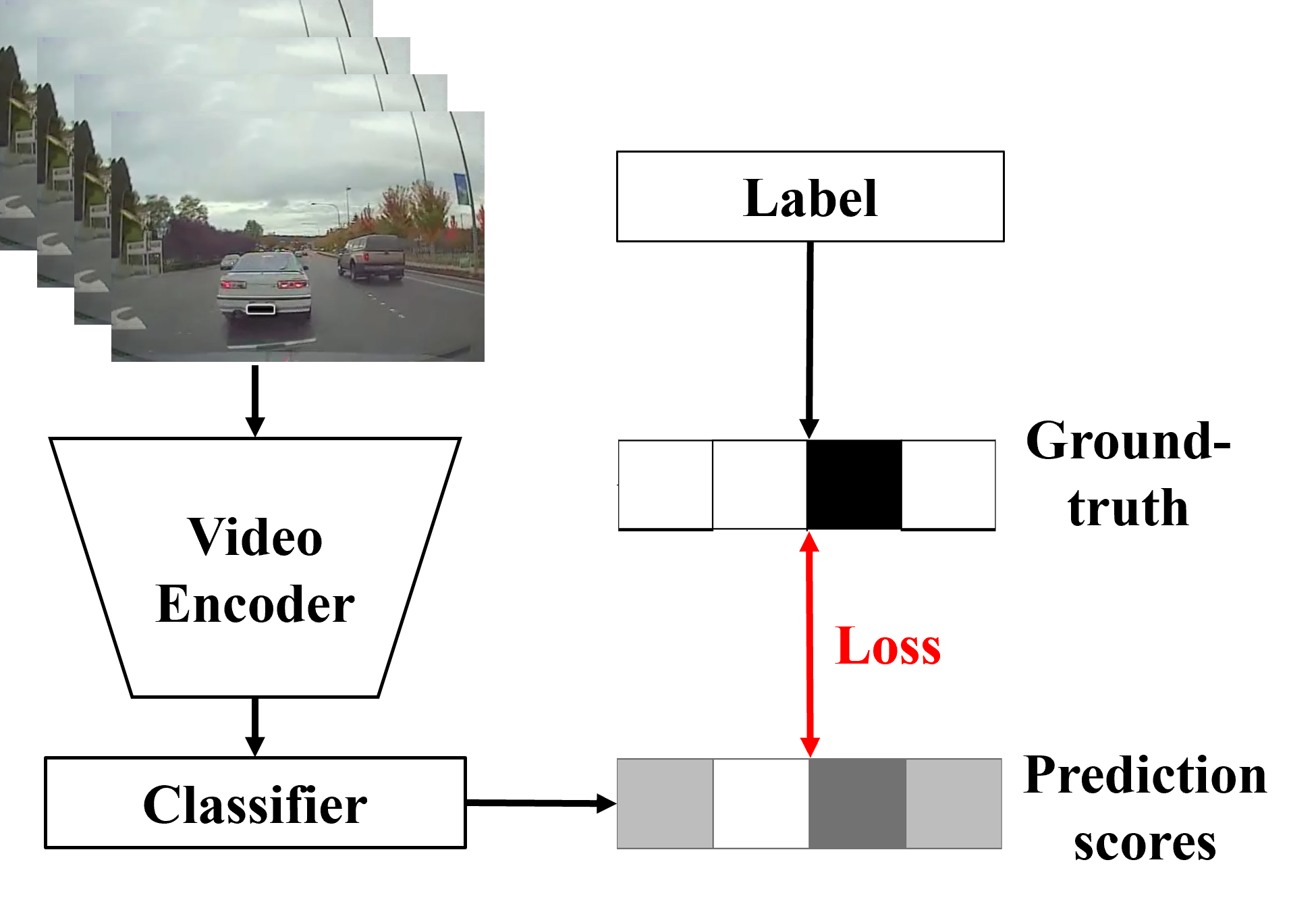}
  \caption{\label{fig:supervised learning structure}Supervised learning structure for video data.}
\end{figure}

Supervised learning for event type classification from video is a 1-of-N vote problem, as illustrated in Figure \ref{fig:supervised learning structure}. This type of model takes a video as input and feeds it through a video encoder to generate video representation. The representation is subsequently processed by a classifier to produce prediction scores. The model is optimized by minimizing the cross entropy loss based on the prediction scores. Given an input video $ x$ and a label $ y$ from a predefined set of labels $ Y $, supervised learning approaches typically estimate the model parameter $\theta$ to compute the conditional probability $ P(y|x, \theta) $. 

The supervised learning approach employs a video encoder $ d_V $, which extracts representations for video data. Then, the classifier projects the video representations into the space with the dimension of labels to obtain the prediction scores:
\begin{equation}\label{1}
\tilde{y_{et}} = \text{Classifier}[d_V(x)] \tag{1}
\end{equation}
Subsequently, the loss to be optimized is defined as the cross-entropy loss between prediction scores and the ground truth:
\begin{equation}\label{2}
\begin{split}
L = \text{Cross Entropy} [ \tilde{y_{et}}, y ]
\end{split}
\tag{2}
\end{equation}
where the ground-truth label $ y $ is converted into a numerical representation or a one-hot vector that indicates its position within the entire label set of length $ |Y| $. During the inference phase, the index with the highest score from the predictions is considered the corresponding category.

\subsection{Contrastive Learning for Conflict Type Classification}

\begin{figure}
  \centering
  \includegraphics[scale=0.45]{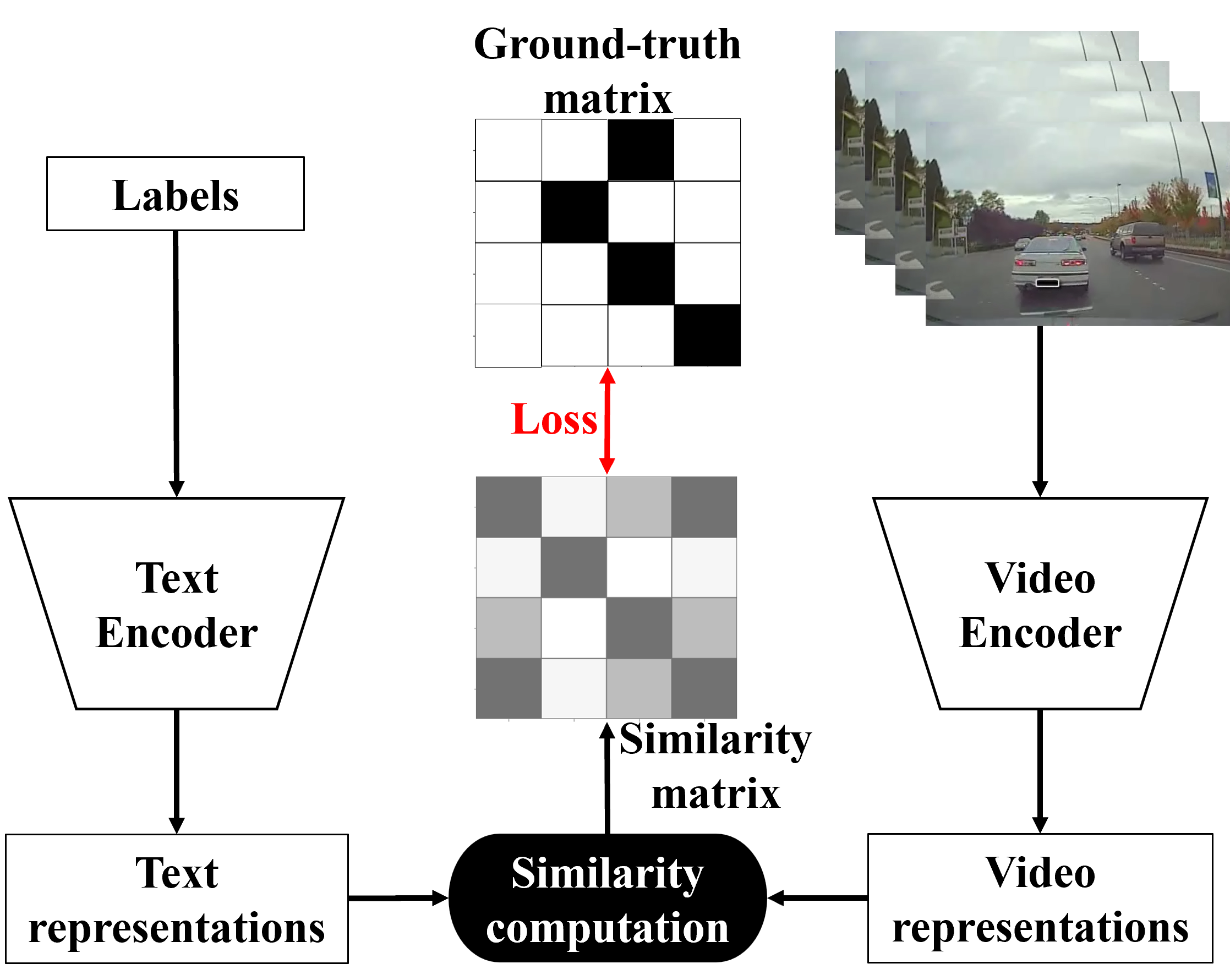}
  \caption{\label{fig:CLIP structure for video data}Contrastive learning structure for video-text pair data.}
\end{figure}

The contrastive learning approach is illustrated in Figure \ref{fig:CLIP structure for video data}. This approach processes a video-text pair as input. The input video is fed into the video encoder to generate video representations. Concurrently, the label text is fed into the text encoder to obtain text representations. The contrastive learning approach computes a similarity score matrix between the video and text representations and is optimized by minimizing the loss between this similarity matrix and the ground-truth video-text pair matrix.

In contrastive learning, the video classification task is redefined as predicting the probability $ P[f(x, y)|\theta] $, where $ y $ represents the original label texts, $\theta$ refers to model parameters, and $ f $ denotes a similarity function. Subsequently, the inference becomes a matching process, with the label texts having the highest similarity score being the outcome:
\begin{equation}\label{3}
  \hat{y_{ct}} = \arg \max_{y \in Y} P[f(x, y)|\theta] \tag{3}
\end{equation}
A contrastive learning approach employs separate encoders $ g_V $ and $ g_T $ for videos and label texts within a dual-stream framework. The video encoder $ g_V $ extracts spatio-and-temporal representations for video data, while the language encoder $ g_T$ captures representations from label texts. To bring matched video and label representations closer, the similarity score is defined using cosine distances:
\begin{equation}\label{4}
s(x, y) = \frac{v^Tt}{\|v\|\|t\|}, \quad s(y, x) = \frac{t^Tv}{\|t\|\|v\|} \tag{4}
\end{equation}
where $ v = g_V(x) $ and $ t = g_T(y) $ represent the encoded representations of $ x $ and $ y $, respectively. Subsequently, the softmax-normalized video-to-text and text-to-video similarity scores are computed as:
\begin{equation}\label{5}
\begin{split}
p_{x \to y}(x) = \text{SoftMax}[s(x, y)]\\
p_{y \to x}(y)= \text{SoftMax}[s(y, x)]
\end{split}
\tag{5}
\end{equation}
The ground-truth similarity scores are denoted as $ q_{x \to y}(x) $ and $ q_{y \to x}(y) $, respectively. The negative pair has a similarity of 0, and the positive pair has a similarity of 1. The video-text contrastive loss to be optimized is defined as 
\begin{equation}\label{6}
\begin{split}
L = \frac{1}{2} \mathbb{E}_{(x, y) \sim D} [ l(p_{x \to y}(x), q_{x \to y}(x)) \\
+ l(p_{y \to x}(y), q_{y \to x}(y)) ]
\end{split}
\tag{6}
\end{equation}
where $ D$ is the training set; $l$ is either cross-entropy loss (for single-label dataset) or Kullback–Leibler (KL) divergence (for multi-label dataset). 

\begin{figure}
  \centering
  \includegraphics[scale=0.45]{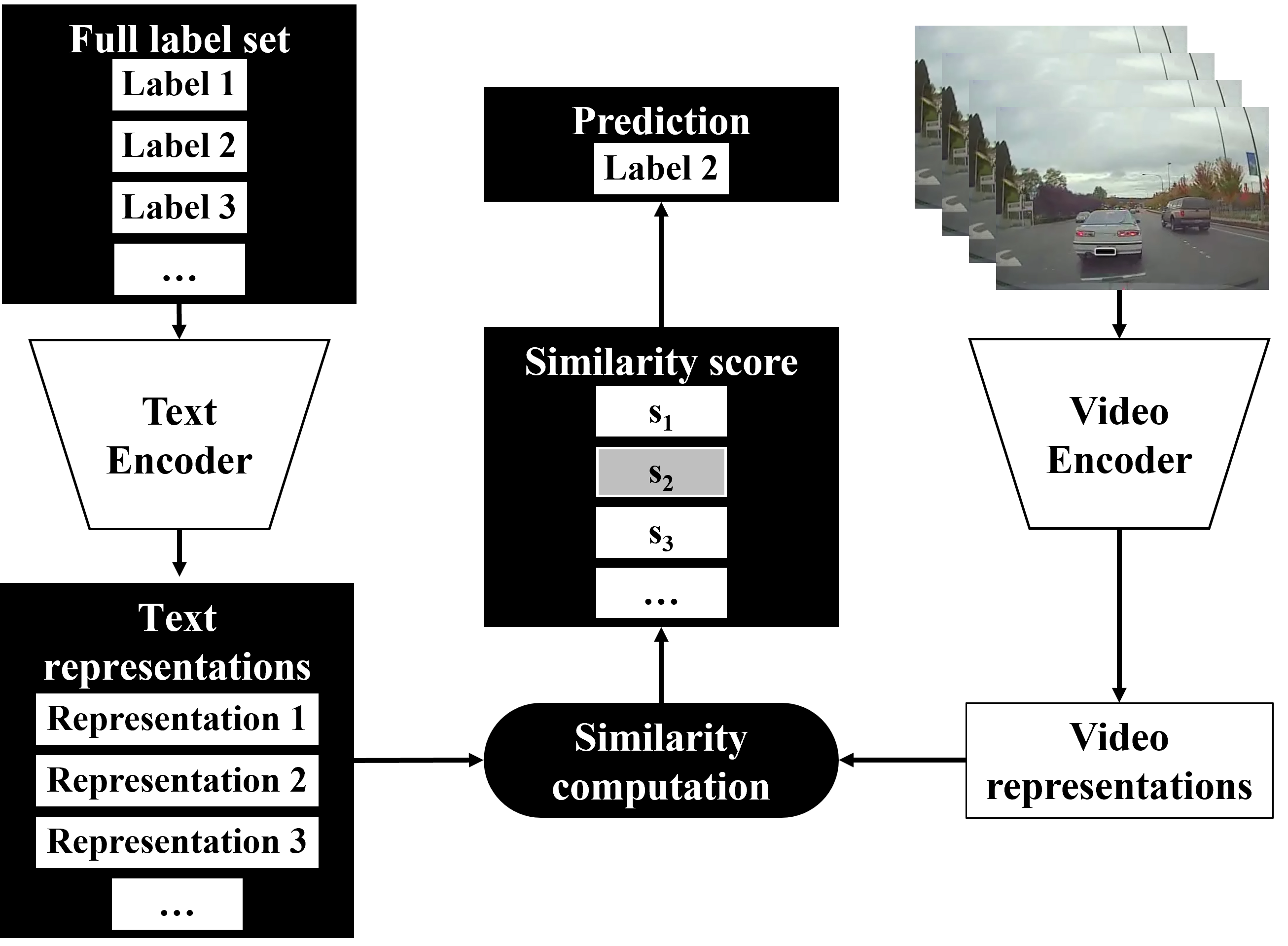}
  \caption{\label{fig:zero shot inference}Inference procedure of contrastive learning approach.}
\end{figure}

A model trained by the contrastive learning approach can carry out inference, as illustrated in Figure \ref{fig:zero shot inference}. When presented with a testing dataset with a label set comprising $M$ labels, the initial step involves extracting the label representations, $ [t_k], k=1,2,..., M$, using the text encoder, $ g_T$. Subsequently, for a given testing video, its representation $ v $ is obtained through the video encoder, $ g_V $. The similarity between $ v $ and each label representation $t_k$ is computed using Equation~\eqref{4}. The label assigned to the video is the one with the highest similarity score with $ v $.

\subsection{Language Models for Event Narrative Generation}

In this study, a VLM is utilized to generate narrative descriptions based on environmental information, such as weather conditions, geographical location, and surrounding context. The process involves the VLM performing inference when provided with a text prompt and the result of SCE detection, enabling accurate event description.

The video representation $ r $  for a given input video $ x$ is obtained using VLM's video encoder: 
\begin{equation}\label{7}
\begin{split}
 r = \text{VLM VideoEncoder}(x)
\end{split}
\tag{7}
\end{equation}
The representation $ r $ is subsequently processed through the Spatial-Temporal Convolution (STC) connector to capture spatial-temporal dynamics. Given a text prompt $ P $, along with the predicted event type $\hat{y_{et}}$ and conflict type $\hat{y_{ct}}$ (if applicable), the output response $R$ is generated from an LLM:
\begin{equation}\label{8}
\begin{split}
R = \text{LLM}[\text{STC}(r), P, \hat{y_{et}}, \hat{y_{ct}}]
\end{split}
\tag{8}
\end{equation}

\section{Application and Results}

\noindent \textbf{Problem Setup}  Utilizing the SHRP 2 NDS dataset \cite{hankey2016description}, this study focused on generating accurate narratives from front-view video of SCEs. In the SHRP 2 data set, normal driving segments were captured a few seconds before SCEs within the trip as a reference level. The dataset includes 1,063 crashes, 774 tire strikes, 6,782 near-crashes, and 8,497 normal driving segments. Each event consists of 30-seconds of front-view video. The SCEs are classified into 16 conflict types, as shown in Table \ref{tab:sce_count}. 

\begin{table}[htpb]
    \centering
    
    \begin{tabular}{lll}
        \toprule
        ID & Conflict type & Count\\ 
        \hline
        1 & Conflict with a lead vehicle & 3165\\
        2 & Single vehicle conflict & 1441\\
        3 & Conflict with vehicle turning into  & \\ 
         & another vehicle path (same direction) & 377 \\
        4 & Conflict with parked vehicle & 173 \\
        5 & Conflict with vehicle in adjacent lane & 1508 \\
        6 & Conflict with vehicle turning across  & \\ 
         & another vehicle path (opposite direction) & 242 \\
        7 & Conflict with a following vehicle & 181 \\
        8 & Conflict with vehicle turning into  & \\ 
         & another vehicle path (opposite direction) & 316 \\
        9 & Conflict with vehicle moving across  &\\ 
         & another vehicle path (through intersection)  & 170 \\
        10 & Conflict with animal & 360 \\
        11 & Conflict with vehicle turning across &\\ 
         & another vehicle path (same direction)  & 65 \\
        12 & Conflict with merging vehicle & 121 \\
        13 & Conflict with pedal cyclist & 64 \\
        14 & Conflict with pedestrian & 163 \\
        15 & Conflict with obstacle/object in roadway & 176 \\
        16 & Conflict with oncoming traffic & 78 \\
        17 & Unknown & 19 \\
        \bottomrule
    \end{tabular}
    \caption{Count of SCEs by conflict types.}
    \label{tab:sce_count}
\end{table}

The proposed approach aims to address three tasks: (1) a classification task to distinguish event types, (2) a classification task to differentiate conflict types, and (3) a text generation task to produce event narratives. To the best of the authors' knowledge, this is the only publicly available driving video dataset with labeled event and conflict types suitable for these three tasks, thereby supporting the evaluation of the proposed approach.

\subsection{SHRP 2 NDS Dataset}

\begin{table*}[htpb]
    \centering
        
        \resizebox{\textwidth}{!}{%
        \begin{tabular}{lccccccc}
            
            \toprule
            Model & Learning Approach & Accuracy & mAP & AUC & Balanced Accuracy & Macro Precsion & Macro F1 \\
            \hline
            X-CLIP& Contrastive & 0.829& 0.708 & 0.937& 0.653&  0.688& 0.666\\
            Action CLIP & Contrastive & 0.816 & 0.659 & 0.901 & 0.639 & 0.646 & 0.642\\
            SlowFast & Supervised & \textbf{0.917} & \textbf{0.862} & \textbf{0.981} & \textbf{0.787} & \textbf{0.811} & \textbf{0.797} \\
            Swin Transformer & Supervised & 0.894 & 0.810 & 0.969 & 0.738 & 0.776 & 0.755\\ 
            TimeSformer & Supervised & 0.851 & 0.727 & 0.950 & 0.650 & 0.691 & 0.668\\
          \bottomrule
        \end{tabular}%
        }
        \caption{Comparison of different models in event type classification.}
        \label{tab:task_1_full}
    \end{table*}

\begin{table*}[htpb]
    \centering
        
        \resizebox{\textwidth}{!}{%
        \begin{tabular}{lcccccccc}
            
            \toprule
            Model \& Training set & Learning Approach & Accuracy & Top5 Acc & mAP & AUC & Balanced Acc & Macro Precsion & Macro F1 \\
            \hline
            X-CLIP (full)& Contrastive & \textbf{0.766}& \textbf{0.951} & \textbf{0.547} & 0.921& \textbf{0.493} & \textbf{0.599} & \textbf{0.508}\\
            Action CLIP (full) & Contrastive & 0.748 & 0.949 & 0.488 & 0.907 & 0.439 & 0.520 &  0.458 \\
            SlowFast (full) & Supervised & 0.721 & 0.928 & 0.467 & \textbf{0.927} & 0.437 & 0.469 & 0.423\\
            Swin Transformer (full) & Supervised & 0.719 & 0.927 & 0.432 & 0.889 & 0.411  & 0.450 & 0.420\\
            TimeSformer (full) & Supervised & 0.713 & 0.945 & 0.468 & 0.920 & 0.448 & 0.487 &0.459\\
            \hline
            X-CLIP (5\%) & Contrastive & \textbf{0.636}& \textbf{0.847}& \textbf{0.278} & 0.770 & \textbf{0.232} & \textbf{0.259} & \textbf{0.222} \\
            Action CLIP (5\%) & Contrastive & 0.606 & 0.846 & 0.239 & 0.777 & 0.216 & 0.244 & 0.220 \\
            SlowFast (5\%) & Supervised & 0.485 & 0.806 & 0.148 & 0.659 & 0.130 & 0.156 & 0.105 \\
            Swin Transformer (5\%) & Supervised & 0.545 & 0.829 & 0.197 & 0.752 & 0.163 & 0.158 & 0.153 \\
            TimeSformer (5\%) & Supervised & 0.571 & 0.840 & 0.205 & \textbf{0.786} & 0.166 & 0.177 &0.154 \\
            \bottomrule
        \end{tabular}%
        }
        \caption{Comparison of different models in conflict type classification.}
        \label{tab:task_2_full}
    \end{table*}

The SHRP 2 NDS is the largest naturalistic driving study to date, involving over 3,000 participants and collecting data from vehicles equipped with a comprehensive data recording system \cite{hankey2016description, dingus2016driver}. This system captured continuous video footage at 15 FPS from four camera angles, resulting in over a million hours, or 70 million miles, of driving data. SCEs, including crashes, tire strikes, and near-crashes, were identified through kinematic data analysis and video verification \cite{hankey2016description}. Near-crashes are defined as situations requiring evasive maneuvers to avoid a crash\cite{hankey2016description}, while tire strikes are linked to road departure incidents \cite{kidd2015relevance}. The extensive dataset and detailed classification of SCEs, available on the SHRP 2 InSight website \cite{shrp2insight}, provide valuable insights into real-world driving behaviors and safety-critical situations.

\vspace{5pt}

\noindent \textbf{Data Pre-processing}  The time of each SCE was pinpointed using the impact timestamp from the SHRP 2 database and serves as the center point of the event \cite{shrp2insight}. A temporal window that  included 38 video frames (equivalent to 2.5 seconds)  both preceding and succeeding the event was extracted, resulting in a 5-second interval of the front-view video. For each SCE, a matched  normal driving segment with the same duration as an SCE was randomly selected from the same trip prior to the SCE.

\subsection{Classification Task Implementation and Performance}

\vspace{5pt}

\noindent \textbf{Model Implementation} The dataset was randomly split into training, testing, and validation subsets in a 7:2:1 ratio. 
Few-shot evaluation utilized 5\% of the conflict type classification training set, with 10 categories containing fewer than 10 samples each. Validation sets were used for hyperparameter tuning, while independent testing sets assessed performance. The environment consisted of Python 3.8 on Rocky Linux 9.3, with model training performed on a workstation equipped with dual Intel Xeon Gold 6338 CPUs, 256 GB RAM, and two Nvidia Tesla A100 80 GB GPUs.

This study evaluated five supervised and contrastive learning approaches to select a suitable method for ScVLM, including X-CLIP \cite{ni2022expanding} and ActionCLIP \cite{wang2023actionclip} for contrastive learning, and SlowFast\cite{Feichtenhofer_2019_ICCV}, Video Swin Transformer \cite{liu2022video}, and TimeSformer \cite{bertasius2021space} for supervised learning. These five models have similar performance on the Kinetics-400 dataset \cite{kinetics400}.

We used the following setup for each model. X-CLIP uses the ViT-B/16 CLIP architecture with a cross-frame communication transformer and a one-layer multi-frame integration transformer. ActionCLIP incorporates six Transformer adapter layers into the ViT-B/16 CLIP architecture. SlowFast employs a ResNet3D backbone, Video Swin Transformer utilizes the Swin-Base architecture, and TimeSformer adopts a TimeSformer-Base model with divided space-time attention. Both TimeSformer and Video Swin Transformer were initialized with ImageNet-22k pre-trained weights. All models were trained with batch sizes optimized for two Tesla A100 GPUs, with the best validation accuracy epoch selected for testing on an independent set.

\begin{table*}[htpb]
    \centering
        
        \resizebox{\textwidth}{!}{%
        \begin{tabular}{lcccccccc}
            
            \toprule
            Model & Learning Approach & Accuracy & Top5 Acc & mAP & AUC & Balanced Acc & Macro Precsion & Macro F1 \\
            \hline
            X-CLIP& Contrastive & \textbf{0.766} & \textbf{0.951} & \textbf{0.547} & 0.921& \textbf{0.493} & \textbf{0.599}& \textbf{0.508} \\
            SlowFast & Supervised & 0.721 & 0.928 & 0.467 & \textbf{0.927} & 0.437 & 0.469 & 0.423\\
            CLIP + mean pooling& Supervised & 0.609& 0.898& 0.286& 0.849& 0.223& 0.317&0.227\\ 
            CLIP + LSTM& Supervised & 0.611& 0.872& 0.243& 0.843& 0.227& 0.212&0.215\\
            \bottomrule
        \end{tabular}%
        }
        \caption{Comparison of alternative models in conflict type classification.}
        \label{tab:task_2_alter}
\end{table*}

\vspace{5pt}

\noindent \textbf{Benchmark Comparison} Six metrics were used to evaluate model performance: Accuracy, mean average precision (mAP), area under the ROC curve (AUC), balanced accuracy, macro precision, and macro F1. The latter three focus on imbalanced scenarios, suitable for the rare-event nature of SCEs \cite{scikit-learn}.

Table \ref{tab:task_1_full} presents the results of four-way event type classification. In general, the supervised learning-based models outperformed the contrastive learning-based models, with SlowFast achieving the best performance across all evaluation metrics. This suggests that on the SHRP 2 NDS dataset, selected supervised learning approaches are more effective for event-type classification task than the selected contrastive learning approaches.

Table \ref{tab:task_2_full} presents a comprehensive comparison for 16-way conflict type classification.  The results include both the full dataset and a limited 5\% training subset for few-shot learning. In the full dataset analysis, contrastive learning-based models outperformed supervised learning-based models across most evaluation metrics, with X-CLIP demonstrating the best overall performance on the SHRP 2 NDS front-view video dataset. 

For few-shot learning, contrastive learning-based models significantly outperformed their supervised learning counterparts across most metrics, with notable improvements in balanced accuracy, macro precision, and macro F1 score. These results indicate that contrastive learning approaches are more effective for conflict type classification than supervised learning approaches on the SHRP 2 NDS dataset, particularly when dealing with minority classes and limited data availability.

\vspace{5pt}

\begin{figure}
  \centering
  \includegraphics[scale=0.45]{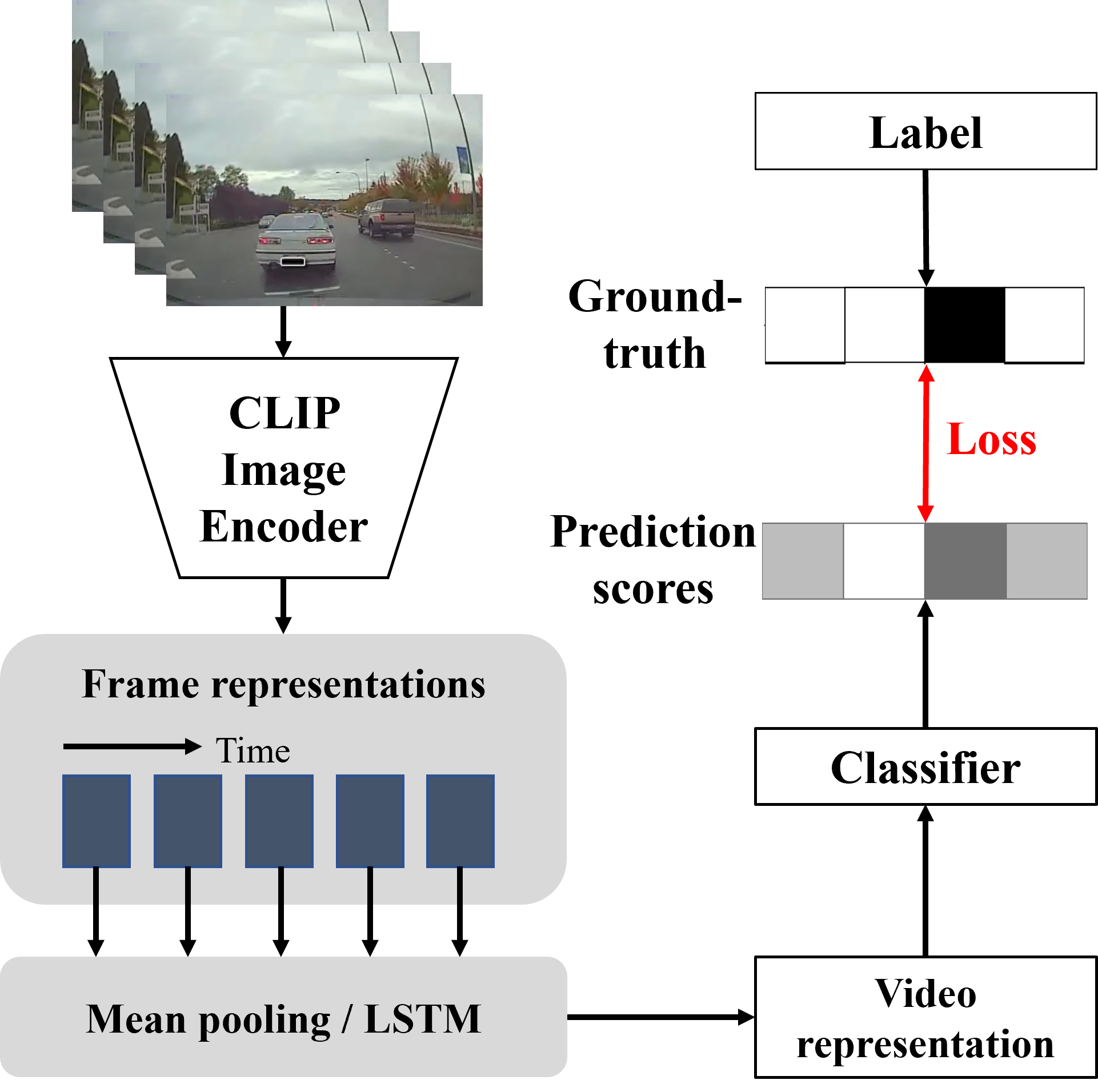}
  \caption{\label{fig:CLIP supervised learning}Conflict type classification with CLIP image encoder in supervised learning approach.}
\end{figure}

\noindent \textbf{Component Performance Evaluation} To evaluate whether the most effective component of contrastive learning approaches in 16-way conflict type classification was the contrastive learning approach or the CLIP encoder, two alternative models were implemented. The raw video frames were processed through a CLIP image encoder, and two methods were employed to handle temporal dependencies across frames: mean pooling and long short-term memory (LSTM) \cite{wang2023actionclip}. The resulting video representations from each method were input into a multi-layer perceptron classifier. The overall process is illustrated in Figure \ref{fig:CLIP supervised learning}.

Table \ref{tab:task_2_alter} compares X-CLIP and SlowFast, the leading models for conflict type classification task using contrastive learning and supervised learning respectively. The performance of the CLIP image encoder in the supervised learning approach was notably lower, suggesting that the CLIP image encoder may not be well-suited for supervised learning approaches in conflict type classification. This evaluation confirms that the superior performance of the CLIP-based contrastive learning approach can be attributed to the model's architecture, with the text encoder playing a crucial role, especially for labels with rich text.

\subsection{Narrative Generation Implementation and Performance} 

\noindent \textbf{Model Implementation} The narrative generation process consists of two steps: environment information extraction and  narrative generation. The VideoLLaMA2 model is employed for understanding environment information using the prompt ``Describe this driving event from dashcam view." We used CLIP ViT-Large-Patch14-336 as the video encoder and Mistral-7B-Instruct-v0.2 as the language decoder \cite{damonlpsg2024videollama2}. 

Narrative generation combines generated environment information with classification results, based on the most effective models: SlowFast for event-type classification and X-CLIP for conflict-type classification. Narrative generation used LLaMA 3.1 8B \cite{dubey2024llama} with the system prompt ``This is related to a driving event. Describe objectively." If the event type was ``Normal Driving," the narrative was generated with the user prompt ``Describe this event: 1: {Environment}. 2: Normal Driving." For SCEs, the narrative was generated using the user prompt ``Describe this event: 1: {Environment}. 2: {Event Type}. 3: {Conflict Type}." To make a fair comparison with other VLMs, the language models were not fine-tuned.

\vspace{5pt}

\begin{figure}[h!]
  \centering
  \includegraphics[scale=0.4]{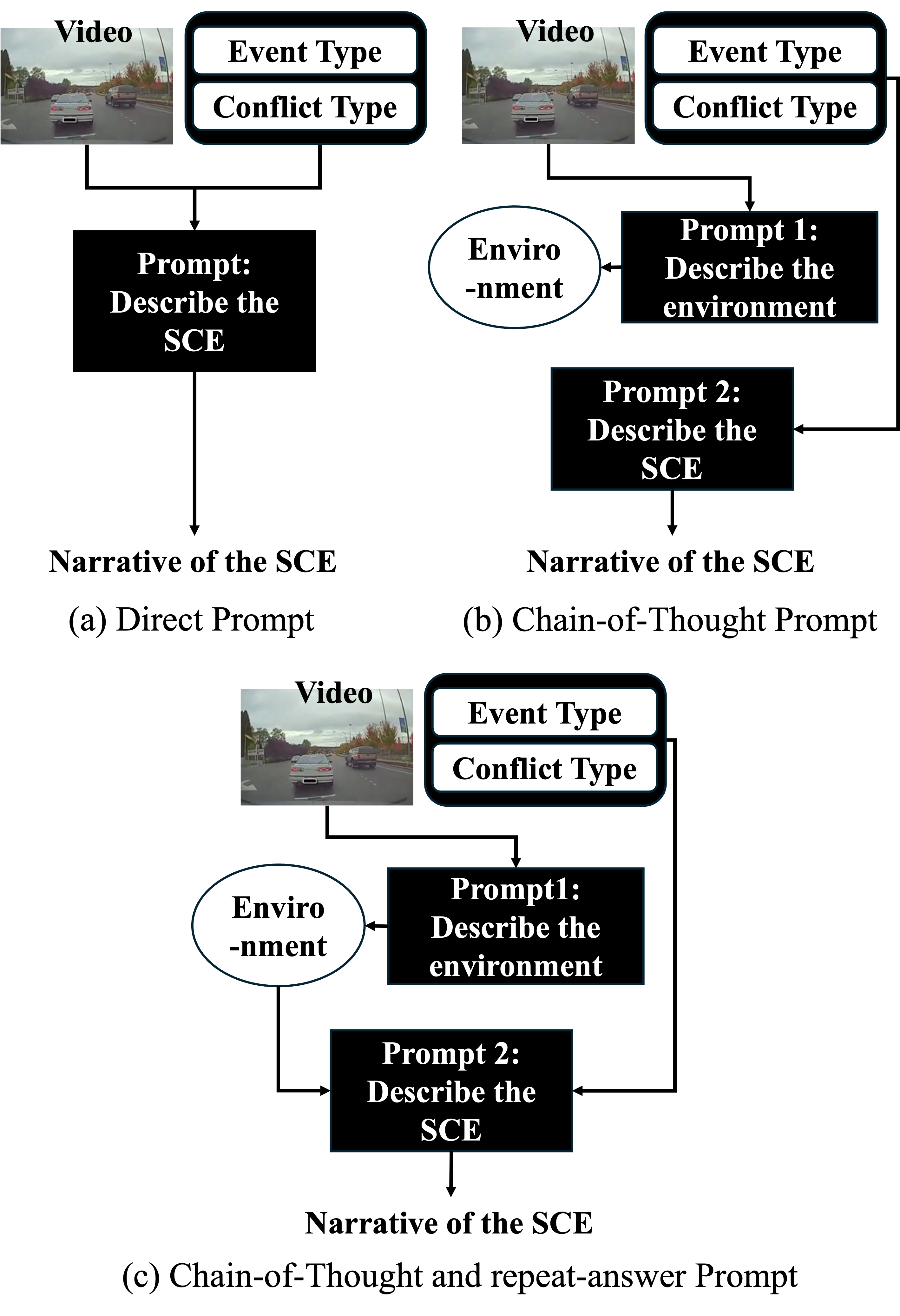}
  \caption{\label{fig:prompt}Prompt strategies.}
\end{figure}

\noindent \textbf{Alternative Prompts for Language Models} Hallucinations in VLMs occur when responses lack factual support or context \cite{favero2024multi}. To mitigate this, this study employs a chain-of-thought prompt \cite{tonmoy2024comprehensive} combined with a repeat-answer \cite{xu2023re} strategy. The VLM first describes the environment, then generates the SCE narrative using event type, conflict type, and environment description. To evaluate this approach, 20 randomly selected SCEs from the testing set were assessed using three strategies: 1) direct prompt 2) chain-of-thought prompt 3) chain-of-thought with repeat-answer (proposed strategy). These are illustrated in Figure \ref{fig:prompt}.

\begin{figure}[h!]
  \centering
  \includegraphics[scale=0.25]{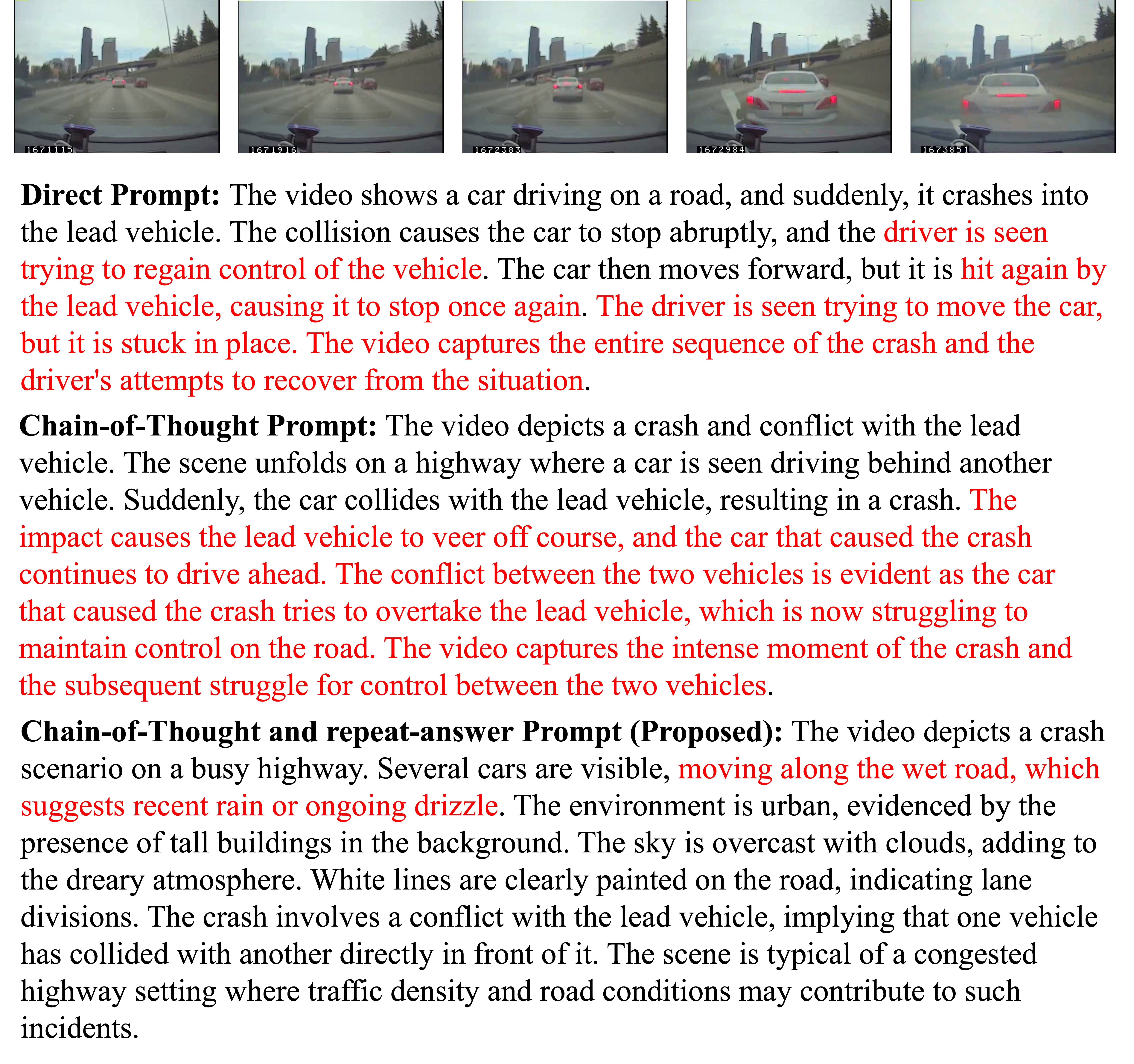}
  \caption{\label{fig:keyframes}Key frames and generated narratives for a lead vehicle crash example (red highlights are the hallucinations).}
\end{figure}

As an example, Figure \ref{fig:keyframes} presents key frames and generated narratives for a lead vehicle crash on a highway, with hallucinations highlighted in red. Among the different strategies, the proposed approach produced the most accurate descriptions, minimizing hallucinations.

\vspace{5pt}

\begin{figure}[h!]
  \centering
  \includegraphics[scale=0.25]{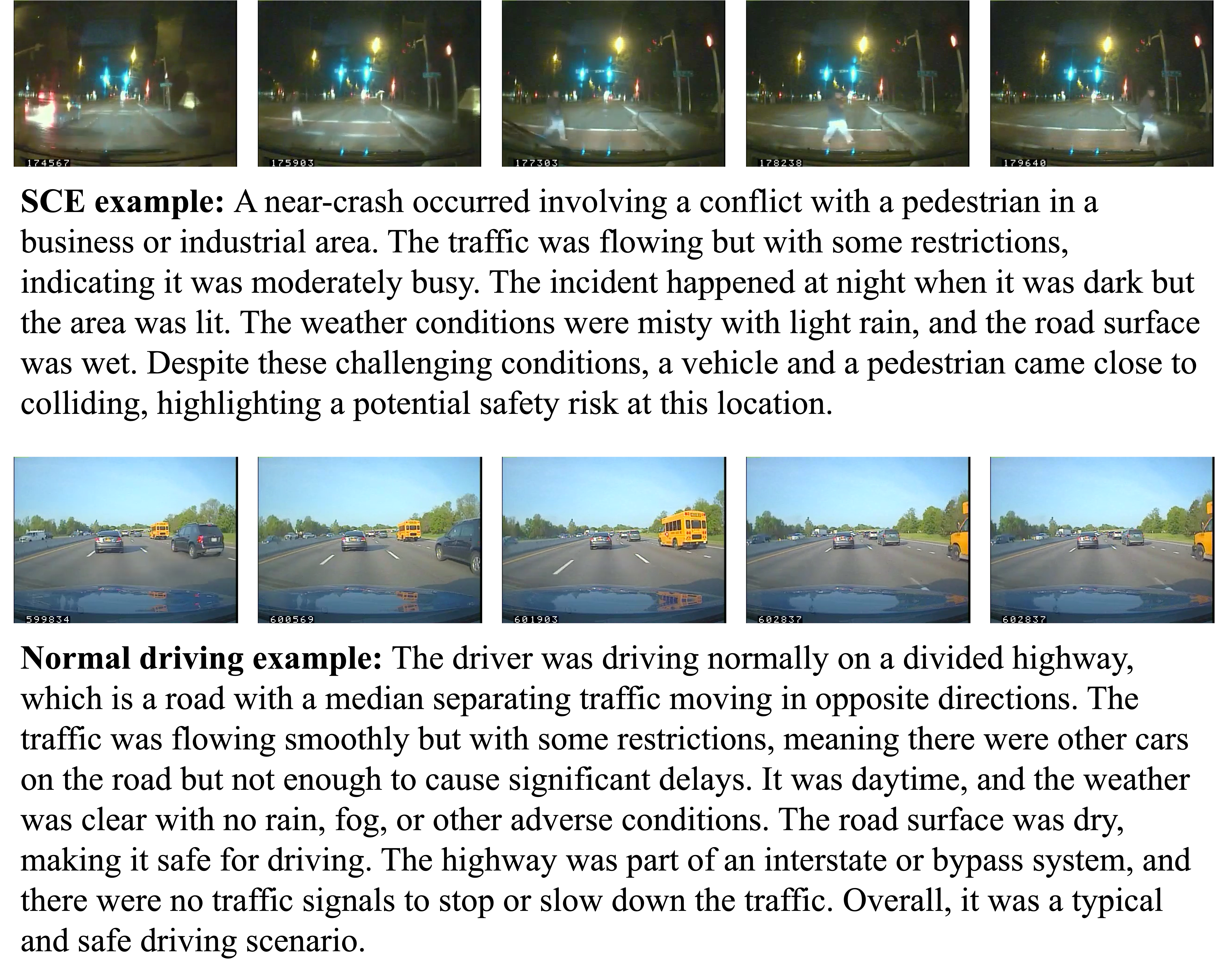}
  \caption{\label{fig:gt_example}Ground truth narratives for normal driving and SCE examples.}
\end{figure}

\noindent \textbf{Benchmark Comparison for Narrative Generation} To evaluate the performance of ScVLM, 100 events were randomly selected, and ground truth narratives were manually annotated by trained volunteers. The annotations include traffic density, light conditions, surface conditions, and locality. For SCEs, additional annotations specify event type, conflict type, and incident type. Two examples are provided, with the key frames shown in Figure \ref{fig:gt_example}.

Among the 100 selected events, 56 were classified as SCEs. The analysis was conducted on two evaluation sets: the full set of selected events and the subset of SCEs. Ten state-of-the-art understanding VLMs were evaluated against ScVLM using metrics ROUGE-L \cite{lin2004rouge}, METEOR \cite{banerjee2005meteor}, and BERTScore \cite{zhang2019bertscore}. The benchmark models used their default setup. For fair comparison, the benchmarks employed a chain-of-thought prompting approach with two sequential prompts: ``Describe this driving event from dashcam view." and ``If there is a safety critical event, describe it." The responses were then combined to form the final narrative. To comprehensively evaluate the generative narratives relative to the ground truth, F1 scores from ROUGE-L and BERTScore were used, providing a balanced measure of both precision and recall.

As shown in Table \ref{tab:narra_comp}, ScVLM outperformed all other models in the full evaluation set, achieving the highest ROUGE-L and BERTScore. This demonstrates that ScVLM excels in narrative generation tasks compared to existing benchmarks. In the SCE subset, ScVLM shows a more pronounced advantage, surpassing all models across every evaluation metric. Specifically, ScVLM outperformed the second-best models by 13.7\% in ROUGE-L, 7.5\% in METEOR, and 5.0\% in BERTScore. This substantial improvement highlights ScVLM's superior performance in SCE narrative generation. ScVLM is the only model to show consistent improvement across all evaluation metrics in the SCE setting. This indicates that ScVLM is particularly robust to the challenges posed by the SCE scenario, outperforming all other models in terms of narrative generation quality.

\begin{table}[htpb]
    \centering
    
    \begin{tabular}{llccc}
        \toprule
        Model\& Eval set & Size & ROUGE-L & METEOR & BERT\\ 
        \hline
        ScVLM (all) & 15B & \textbf{0.186} & 0.197 & \textbf{0.572} \\
        VideoLLaMA 2 \cite{damonlpsg2024videollama2} (all) & 7B & 0.165 & 0.119 & 0.543 \\
        Qwen2-VL \cite{wang2024qwen2} (all) & 72B & 0.173 & 0.173 & 0.559 \\
        MiniCPM-V \cite{yao2024minicpm} (all) & 8B & 0.169 & 0.179 & 0.563 \\
        LLaVA-Next-Video \cite{zhang2024llavanextvideo} (all) & 34B & 0.138 & \textbf{0.204} & 0.542 \\
        LLaVA-OneVision \cite{li2024llava} (all) & 72B & 0.159 & 0.163 & 0.525 \\
        Chat-UniVi \cite{jin2024chat} (all) & 13B & 0.162 & 0.199 & 0.559 \\
        LLaMA-Vid \cite{li2025llama} (all) & 13B & 0.178 & 0.142 & 0.536 \\
        PLLaVA \cite{xu2024pllava} (all) & 13B & 0.130 & 0.189 & 0.530 \\
        Video-ChatGPT \cite{maaz2023video} (all) & 7B & 0.132 & 0.145 & 0.521 \\
        Video-LLaVA \cite{lin2023video} (all) & 7B & 0.161 & 0.167 & 0.541 \\
        \hline
        ScVLM (SCE) & 15B & \textbf{0.199} $\uparrow$  & \textbf{0.215} $\uparrow$ & \textbf{0.589} $\uparrow$\\ 
        VideoLLaMA 2 \cite{damonlpsg2024videollama2} (SCE) & 7B & 0.163 $\downarrow$ & 0.120 $\uparrow$ & 0.548 $\uparrow$ \\
          Qwen2-VL \cite{wang2024qwen2} (SCE) & 72B & 0.169 $\downarrow$ & 0.169 $\downarrow$ & 0.555 $\downarrow$ \\
        MiniCPM-V \cite{yao2024minicpm} (SCE) & 8B & 0.168 $\downarrow$ & 0.179 $-$ & 0.558 $\downarrow$ \\
        LLaVA-Next-Video \cite{zhang2024llavanextvideo} (SCE) & 34B & 0.135 $\downarrow$ & 0.200 $\downarrow$ & 0.533 $\downarrow$ \\
        LLaVA-OneVision \cite{li2024llava} (SCE) & 72B & 0.157 $\downarrow$ & 0.166 $\uparrow$ & 0.525 $-$\\
        Chat-UniVi \cite{jin2024chat} (SCE) & 13B & 0.157 $\downarrow$ & 0.194 $\downarrow$ & 0.561 $\uparrow$ \\
        LLaMA-Vid \cite{li2025llama} (SCE) & 13B & 0.175 $\downarrow$ & 0.145 $\uparrow$ & 0.545 $\uparrow$ \\
        PLLaVA \cite{xu2024pllava} (SCE) & 13B & 0.122 $\downarrow$ & 0.177 $\downarrow$ & 0.522 $\downarrow$ \\
        Video-ChatGPT \cite{maaz2023video} (SCE) & 7B & 0.121 $\downarrow$ & 0.134 $\downarrow$ & 0.520 $\downarrow$ \\
        Video-LLaVA \cite{lin2023video} (SCE) & 7B & 0.154 $\downarrow$ & 0.161 $\downarrow$ & 0.538 $\downarrow$ \\
        \bottomrule
    \end{tabular}
    \caption{Comparison of different models in narrative generation.}
    \label{tab:narra_comp}
\end{table}

\section{Conclusion}

This study introduced ScVLM, an approach that integrates supervised learning, contrastive learning, and VLM. The approach enhances the understanding of driving videos, improves the rationality of event descriptions, and reduces hallucinations in VLM-generated outputs.

Based on the SHRP 2 NDS video dataset, the results demonstrate that the proposed ScVLM generates more precise and contextually appropriate event descriptions compared to a standard VLM. This work not only contributes to the accuracy of SCE detection, but also offers a robust framework for future research in automatic generation of SCE descriptions.

\section*{Acknowledgement}

This project is partially funded by a grant from the National Surface Transportation Safety Center for Excellence (Grant Number: 238717).

\bibliographystyle{unsrtnat}  
\bibliography{references}

\end{document}